\documentclass[letterpaper, 10 pt, conference]{ieeeconf} 
\IEEEoverridecommandlockouts

\usepackage{cite}
\usepackage{amsmath,amssymb,amsfonts}
\usepackage{algorithmic}
\usepackage{graphicx}
\usepackage{textcomp}
\usepackage{xcolor}
\def\BibTeX{{\rm B\kern-.05em{\sc i\kern-.025em b}\kern-.08em
    T\kern-.1667em\lower.7ex\hbox{E}\kern-.125emX}}

\usepackage{times}

\usepackage{multicol}
\usepackage[bookmarks=true]{hyperref}

\usepackage{bm}

\usepackage{amsthm}
\usepackage{url}
\usepackage{svg}
\usepackage{nicefrac}
\usepackage{balance}
\newtheoremstyle{exampstyle}
  {3pt} 
  {3pt} 
  {\itshape} 
  {} 
  {\bfseries} 
  {.} 
  {.5em} 
  {} 

\theoremstyle{exampstyle} 
\newtheorem{definition}{Definition}

\newtheorem{theorem}{Theorem}
\newtheorem{remark}{Remark}
\newtheorem{assumption}{Assumption}
\newtheorem{problem}{Problem}
\newtheorem{proposition}{Proposition}
\newtheorem{example}{Example}
\newtheorem{example*}{Example*}

\theoremstyle{plain}

\DeclareMathOperator*{\argmin}{arg\,min}
\newcommand{\RNum}[1]{\uppercase\expandafter{\romannumeral #1\relax}}

\begin{document}

\renewcommand{\baselinestretch}{0.99}
\newcommand{\high}[1]{{\color{black}#1}}

\title{Belief Control Barrier Functions for Risk-aware Control}

\author{Matti Vahs, Christian Pek and Jana Tumova
	\thanks{This work was partially supported by the Wallenberg AI, Autonomous
		Systems and Software Program (WASP) funded by the Knut and Alice
		Wallenberg Foundation.
  This research has been carried out as part of the Vinnova Competence Center for Trustworthy Edge Computing Systems and Applications at KTH Royal Institute of Technology.}
	\thanks{Matti Vahs and Jana Tumova are with the Division of Robotics, Perception and Learning, KTH Royal Institute of Technology, Stockholm, Sweden and also affiliated with Digital Futures. Mail addresses: {\{\tt\small vahs, tumova\}}
		{\tt\small @kth.se}. Christian Pek is with the Department of Cognitive Robotics (CoR) at Delft University of Technology (TUD). Mail address: {\tt\small C.Pek@tudelft.nl}}%
}

\maketitle

\begin{abstract}
Ensuring safety in real-world robotic systems is often challenging due to unmodeled disturbances and noisy sensor measurements.
To account for such stochastic uncertainties, many robotic systems leverage probabilistic state estimators such as Kalman filters to obtain a robot's \emph{belief}, i.e. a probability distribution over possible states.
We propose belief control barrier functions (BCBFs) to enable risk-aware control synthesis, leveraging all information provided by state estimators.
This allows robots to stay in predefined safety regions with desired confidence under these stochastic uncertainties.
BCBFs are general and can be applied to a variety of robotic systems that use extended Kalman filters as state estimator. 
We demonstrate BCBFs on a quadrotor that is exposed to external disturbances and varying sensing conditions.
Our results show improved safety compared to traditional state-based approaches while allowing control frequencies of up to 1kHz.
\end{abstract}

\section{Introduction}
\begin{figure}[t]
    \centering
    \includegraphics[width=0.35\textwidth]{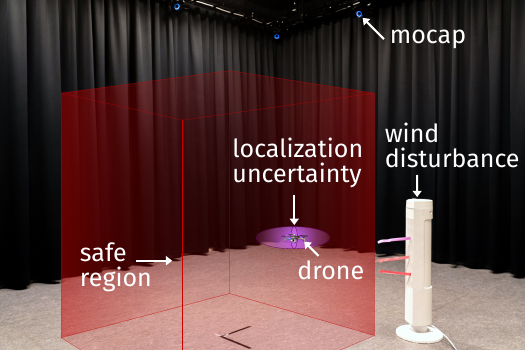}
    \caption{An illustration of our experiments with a CrazyFlie 2.1 quadrotor operating in an uncertain environment. The drone is exposed to common real-world uncertainties such as noisy position measurements and unknown wind disturbances. The uncertainty in the state estimate is shown as a purple ellipsoid and the cuboid-shaped safe set is colored in red.}
    \vspace{-0.5cm}
    \label{fig:RealworldExp}
\end{figure}

Autonomous robotic systems are exposed to various sources of uncertainty, such as noise in the robot's sensor readings or external disturbances, e.g., an unknown wind force acting on a quadrotor, as shown  in Fig. \ref{fig:RealworldExp}. Considering uncertainty is thus crucial for safe operation of such robots. 
Safe and robust control synthesis under external disturbances has been thoroughly explored for a variety of robots with full state information~\cite{ames2019control}. Observer-based Control Barrier Functions~(CBFs) consider measurement uncertainties through bounded estimation errors \cite{Zhang_2022, Wang_2021, Dean_20}. However, these methods do not consider the stochastic uncertainties that are commonly captured in probabilistic state estimation techniques deployed in practice, such as Kalman Filters (KFs). 
Such stochastic uncertainties have been considered for instance in the state-of-the-art chance-constrained nonlinear Model Predictive Control (MPC) \cite{8613928}, which  bounds the probability of undersired events. However, to make the method computationally tractable, theoretical safety guarantees are sacrificed.
In this paper, we consider both uncertainties in the robot's motion and its observations. We aim to offer both safety guarantees under uncertainty as well as a practical solution. 

In practice, state estimation pipelines are used to obtain a robot's belief that accounts for the uncertainty arising from stochastic motion and observations. 
Up to now, the Kalman filter (KF) is still one of the gold standards in robotic state estimation because of its simplicity and robustness \cite{thrun2005probabilistic}. The extended KF (EKF), as its extension to nonlinear systems, is widely used for state estimation of robots such as quadrotors, legged robots or autonomous underwater vehicles~\cite{urrea2021kalman}. One of the key properties of KFs is that they not only provide an estimate about the robot's state but also quantify the uncertainty of that estimate through the covariance matrix. 
For instance, Fig. \ref{fig:RealworldExp} illustrates a drone's localization uncertainty as a Gaussian level set (purple ellipsoid), i.e. a set containing the state with certain probability. In this work, we leverage both the mean state estimate and the covariance matrix.

Due to Gaussian state uncertainty, hard safety constraints on the system states are generally infeasible.
This gives rise to a risk-aware perspective on safety: We consider specifications that bound the probability of violating safety constraints on the state. Fig. \ref{fig:RealworldExp} shows a scenario in which a safety specification is encoded as ``\emph{the probability of leaving the safe region (red cuboid) should be less than 1\%}''.

However, designing controllers that ensure satisfaction of such risk-aware safety specifications is challenging as it requires us to reason about the robot's belief instead of its state. Belief spaces in these applications suffer from the curse of dimensionality and are hybrid in nature as robots evolve in continuous time while sensors only provide measurements at discrete timesteps.
To enable risk-aware control, we introduce Belief CBFs (BCBFs) that
    1) serve as safety filters in the presence of real-world stochastic uncertainties,
    2) provide theoretical safety guarantees under the hybrid nature of belief spaces, and
    3) overcome the curse of dimensionality and allow real-time control for general robotic systems that employ EKFs as a state estimator.
We evaluate our approach in experiments with a quadrotor that is exposed to external wind disturbances and varying sensing conditions, see Fig. \ref{fig:RealworldExp}.

 \subsection{Related Work} 

 We review two relevant safety-critical control approaches for robotic systems under uncertainties -- CBFs and MPC.

 Related work on observer-based CBFs assume a bounded uncertainty around the current estimate that is used to account for potential measurement errors
 \cite{Zhang_2022}, \cite{Wang_2021}.
 However, these methods do not consider stochastic uncertainties that are commonly used in probabilistic state estimators. Measurement-robust CBFs use deterministic measurements and assume a given mapping from measurements to state estimates \cite{Dean_20}. In contrast, our BCBFs explicitly provide this mapping to handle stochastic measurements. 


Control of stochastic systems using Kalman Filters as state estimator has been addressed in various settings \cite{8613928,castillo2020real}. 
Chance-constrained nonlinear MPC (CCNMPC) \cite{8613928} enforce obstacle avoidance with a desired confidence level in which the uncertainty in the state estimate originates from an unscented KF.
Unfortunately, solving NMPC problems in belief spaces is computationally expensive due to the curse of dimensionality.
The problem was made tractable by neglecting the dynamics of the covariance through linearization around the last trajectory solution \cite{8613928}. 
However, this linearization is only valid when the subsequent trajectory does not significantly change.
In contrast, the belief space planning approach SACBP  operates in continuous-time through sequential action control (SAC) \cite{nishimura2021sacbp}. The authors model the belief dynamics as a hybrid dynamical system which we build upon. However, SACBP cannot ensure safety as constraints can only be included in the objective function.

In \cite{8814901}, invariance properties of deterministic CBFs are extended to systems that are described by stochastic differential equations (SDEs). The proposed stochastic CBFs handle uncertainty in the system's dynamics and also in the measurements. 
They guarantee safety under stochastic uncertainties by bounding the estimation errors in an EKF.
This CBF is used for systems under sensor faults and attacks in \cite{9303766} as well as risk-bounded control in highway scenarios in \cite{Yaghoubi21}. 
Probabilistic safety barrier certificates (PrSBC) have been proposed in \cite{NEURIPS2020_03793ef7} for multi-robot collision avoidance under uncertainty.  
However, the guarantees only hold for bounded uniform additive noise on the system dynamics and the observation model. 
Furthermore, the guarantees in \cite{8814901, 9303766, Yaghoubi21, NEURIPS2020_03793ef7} only hold for continuous-time observations, which is generally not consistent with real-world robotic systems, e.g., when global positioning data is not available.

\section{Preliminaries}
\subsection{Control Barrier Functions}
Consider a dynamical system in control affine form
\begin{align}
    \dot{\bm{x}} = \bm{f}\left(\bm{x}\right) + \bm{g}\left(\bm{x}\right) \bm{u}\label{eq:deterministic}
\end{align}
with state $\bm{x} \in \mathcal{X} \subseteq \mathbb{R}^n$ and control input $\bm{u} \in \mathcal{U} \subseteq \mathbb{R}^m$. 
A safe set $\mathcal{C}$ is constructed as the superlevel set of a continuously differentiable function $h: \mathcal{X} \rightarrow \mathbb{R}$ such that
\begin{equation}
\begin{aligned}
  \label{eq:safeset_state}
    \mathcal{C} &= \left\{\bm{x} \in \mathcal{X} \mid h\left(\bm{x}\right) \geq 0\right\},\\
    \partial \mathcal{C} &= \left\{\bm{x} \in \mathcal{X} \mid h\left(\bm{x}\right) = 0\right\}.
 \end{aligned}
\end{equation} 
\begin{definition}
A safe set $\mathcal{C}$ is forward invariant with respect to the system \eqref{eq:deterministic} if for every initial condition $\bm{x}(t_0) \in \mathcal{C}$
it holds that $\bm{x}(t) \in \mathcal{C}, \forall t \geq t_0$.
\end{definition}
A prominent approach to render a safe set forward invariant is to use CBFs.
\begin{definition}
\label{def:CBF}
Given a set $\mathcal{C}$\high{, defined by Eq. \eqref{eq:safeset_state}, $h$ serves as a zeroing CBF for the system \eqref{eq:deterministic} if  $\forall \bm{x}$ satisfying $h\left(\bm{x}\right) \geq 0, \exists \bm{u} \in \mathcal{U}$ such that}
\begin{align}
    \frac{\partial h}{\partial \bm{x}} \left(\bm{f}\left(\bm{x}\right) + \bm{g}\left(\bm{x}\right) \bm{u}\right) \geq - h\left(\bm{x}\right).\label{eq:cbf}
\end{align}
\end{definition}
In case a valid CBF exists, it follows that a controller satisfying Eq.~\eqref{eq:cbf} renders $\mathcal{C}$ forward invariant \cite{ames2016control}. However, this only holds for systems with relative degree $r_b = 1$ \high{because $(\nicefrac{\partial h}{\partial \bm{x}}) \bm{g} = \bm{0}$ otherwise, thus $\bm{u}$ would not appear in Eq.~\eqref{eq:cbf} \cite{7524935}}. For systems with $r_b > 1$, exponential CBFs (ECBFs) can be used to ensure forward invariance of $\mathcal{C}$. We leverage second order CBFs as a special case of ECBFs.
\begin{definition}
\label{def:ECBF}
Given a set $\mathcal{C}$ \high{in Eq. ~\eqref{eq:safeset_state} and a system \eqref{eq:deterministic}} with relative degree $r_b=2$, $h$ is called an exponential CBF, if there exists a gain vector $\bm{\zeta} \in \mathbb{R}^{r_b}$ and \high{$\bm{u} \in \mathcal{U}$} such that
\begin{align}
   \ddot{h}\left(\bm{x}\right) + \bm{\zeta}^T \begin{bmatrix}
    h\left(\bm{x}\right) & \dot{h}\left(\bm{x}\right)
    \end{bmatrix}^T\geq 0.\label{eq:SecondOrderCBF}
\end{align}
\end{definition}
The gain vector can be obtained using classical tools from control theory such as, e.g., pole placement. A proof of forward invariance under ECBFs can be found in \cite{7524935}.

\subsection{Gaussian Belief States}
Gaussian filters are a family of state estimators that describe a Bayesian approach in which the belief is constrained to follow a multivariate Gaussian (MVG). Its probability density function (pdf) is described by
\begin{align}
    p\left(\bm{x}(t)\right) = \mathcal{N}\left(\bm{\mu}(t), \bm{\Sigma}(t)\right), \label{eq:MVG}
\end{align}
where $\bm{\mu} \in \mathbb{R}^n$ is the mean vector and $\bm{\Sigma}=\bm{\Sigma}^T \in \mathbb{R}^{n \times n}$ is the positive semidefinite covariance matrix. The pdf in \eqref{eq:MVG} is uniquely described by the belief state $\bm{b} = \left[\bm{\mu}, \mathrm{vec}\left(\bm{\Sigma}\right)\right]^T$ where, due to symmetry, only the upper triangular matrix is stored which is encoded in the $\mathrm{vec}(\cdot)$ operator \cite{van2012motion}. Thus, the dimensionality of the belief state $n_b$ increases quadratically with the state dimension $n$, i.e. $n_b = \nicefrac{(n^2 + 3n)}{2}$. \high{If not mentioned explicitly otherwise, we refer to the belief state $\bm{b}$ as the belief.}

\subsection{Chance Constraints and Risk Measures}
Chance constraints handle safety constraints under uncertainty by bounding the probability of undesired events. We use safety specifications in the form of half-spaces $\bm{\alpha}^T \bm{x} \geq \beta$. Consider a Gaussian distributed random variable $\bm{x}$ with belief state $\bm{b}$. We calculate the probability of satisfying a half-space constraint as \cite{8613928}
\begin{align}
    \mathrm{Pr}\left[\bm{\alpha}^T \bm{x} \geq \beta\right] = \frac{1}{2} \left(1 - \mathrm{erf}\left(\frac{\bm{\alpha}^T \bm{\mu} - \beta}{\sqrt{2 \bm{\alpha}^T \bm{\Sigma} \bm{\alpha}}}\right)\right), \label{eq:ProbHalfspace}
\end{align}
where $\mathrm{erf}(\cdot)$ is the standard error function. 

To quantify the outcome of violating a chance constraint $\mathrm{Pr}\left[\bm{\alpha}^T \bm{x} \geq \beta\right] \geq 1 - \delta$, we use the Value-at-Risk ($\mathrm{VaR}$).
\begin{definition}
The $\mathrm{VaR}$ of a random variable $x \in \mathbb{R}$ with pdf $p\left(x\right)$ at level $\delta \in (0, 1]$ is the $\left(1 - \delta\right)$ quantile, i.e.
\begin{align}
    \mathrm{VaR}_{\delta}\left(x\right) &= \underset{\tau \in \mathbb{R}}{\mathrm{inf}} \left\{\tau \mid \mathrm{Pr} \left[x \geq \tau\right] \geq 1 - \delta\right\}
\end{align}
\end{definition}
$\mathrm{VaR}$ allows us to formulate constraints that are qualitatively equivalent to chance constraints, i.e. 
\begin{align}\mathrm{Pr}\left[\bm{\alpha}^T \bm{x} \geq \beta\right] \geq 1 - \delta \Leftrightarrow \mathrm{VaR}_{\delta}(\bm{\alpha}^T \bm{x} - \beta) \geq 0.
\label{eq:qual}
\end{align}


\section{Problem Setting}
We consider the robot's stochastic motion and observations in the form
\begin{align}
    \dot{\bm{x}} &= \bm{f}\left(\bm{x}\right) + \bm{g}\left(\bm{x}\right) \bm{u} + \bm{w}, \hspace{0.5cm} \bm{w} \sim \mathcal{N}\left(\bm{0}, \bm{Q}\right)\label{eq:motion}\\
    \bm{z}_k &= \bm{\ell}\left(\bm{x}_k\right) + \bm{v}_k, \hspace{0.5cm} \bm{v}_k \sim \mathcal{N}\left(\bm{0}, \bm{R}_k\right)\label{eq:observation}
\end{align}
where $\mathcal{Z} \subseteq \mathbb{R}^{\ell}$ is the observation space and $\bm{w}, \bm{v}$ are the motion and observation noise, respectively. 
We model the robot's motion  as a continuous-time differential equation. 
The observations are always provided in discrete time due to the sensor's sampling time. 
Especially for exteroceptive sensors like GPS, measurements occur much less frequently than the robot's control rate, encouraging us to consider discrete-time formulations. \high{A reference controller, e.g. a controller that drives the robot to a goal state, is given as~$\bm{u}_{\text{ref}}$.}
\begin{problem}
\label{prob:State}
Given the stochastic dynamics in \eqref{eq:motion}, a reference controller $\bm{u}_{\text{ref}}$, a safe set $\mathcal{C}_x = \left\{\bm{x} \in \mathcal{X} \mid \bm{\alpha}^T \bm{x} \geq \beta\right\}$ defined in state space and a confidence level $\delta \in (0,1]$, synthesize a control \high{law $\bm{u}$ that maps $\bm{x}\times\bm{u}_{\text{ref}}\mapsto \bm{u}$} such that \high{at any time} $\mathrm{Pr}\left[\bm{x}\high{(t)} \in \mathcal{C}_x\right] \geq 1 - \delta$.
\end{problem}
 Ideally, to solve Problem \ref{prob:State}, we would use the Bayes filter to capture the exact time evolution of the belief given an initial belief $p\left(\bm{x}(t_0)\right)$, controls $\bm{u}$ and observations $\bm{z}$. However, exact belief calculations only exist in specialized cases which is why approximations need to be considered \cite{thrun2005probabilistic}. 
 Thus, we use an EKF as a tractable implementation of the Bayes filter in which beliefs are Gaussian. \high{While the Gaussian belief of an EKF is exact for linear systems, it is only an approximation of the true belief in the nonlinear case. This assumption, however, is common in many practical scenarios \cite{urrea2021kalman}, especially if the true probability distribution is unimodal \cite{thrun2005probabilistic}. In future work, we aim to consider the mismatch between the modeled belief and the true belief.}

 To approach Problem \ref{prob:State}, we reason about Gaussian belief states $\bm{b}$ instead of states $\bm{x}$ so that we can solve a relaxed problem under stochastic uncertainties.
We propagate the belief through the nonlinear model in Eq. \eqref{eq:motion}-\eqref{eq:observation} by exploiting the fundamental EKF step that systems are linearized around the current mean.
\begin{assumption}
\label{assumption:linearizable}
To propagate the mean and covariance of a random variable $\bm{x}$ through a nonlinear function $\bm{\eta}$, we use a first-order Taylor series expansion
\begin{align*}
    \mathbb{E}\left\{\bm{\eta}\left(\bm{x}\right)\right\} &\approx \bm{\eta}\left(\mathbb{E}\left\{\bm{x}\right\}\right)\\
    \mathrm{Var}\left\{\bm{\eta}\left(\bm{x}\right)\right\} &\approx \left(\frac{\partial\bm{\eta}}{\partial\bm{x}}\left(\mathbb{E}\left\{\bm{x}\right\}\right)\right)\mathrm{Var}\left\{\bm{x}\right\}\left(\frac{\partial\bm{\eta}}{\partial\bm{x}}\left(\mathbb{E}\left\{\bm{x}\right\}\right)\right)^T.
\end{align*}
\end{assumption}


We translate the safety specification over states in Problem~\ref{prob:State} to a safe set over belief states 
\begin{align}
    \mathcal{C}_b = \left\{\bm{b} \in \mathbb{R}^{n_b} \mid 
   h_b(\bm{b}) \geq 0 \right\}, \label{eq:safeset}
\end{align}
where $h_b$ defines a risk-aware half-space
\begin{equation}
\begin{aligned}
    h_b\left(\bm{b}\right) &:=\mathrm{VaR}_{\delta}\left(\bm{\alpha}^T \bm{x} - \beta\right)\\
    &=\bm{\alpha}^T \bm{\mu} - \beta - \mathrm{erf}^{-1} \left(1 - 2 \delta\right)\sqrt{2 \bm{\alpha}^T\bm{\Sigma} \bm{\alpha}}.
\end{aligned}
\label{eq:halfspace}
\end{equation}
\begin{figure}[t]
\centering
\input{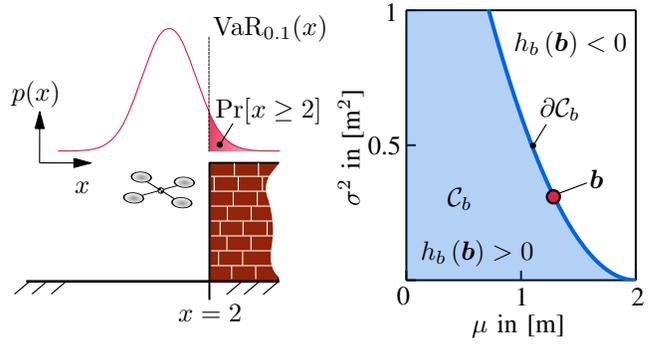}
\caption{A drone with uncertain position $x$ and its pdf $p(x)$ is moving in one dimension. A safety specification is defined as a bounded probability of collision with the wall. The resulting safe set over belief states is shown in blue and the current belief of the robot is depicted in red.}
\vspace{-0.6cm}
\label{fig:Example}
\end{figure}

\renewcommand\theexample{\unskip}
\begin{problem}
\label{prob:Belief}
Given the model \eqref{eq:motion}-\eqref{eq:observation} and a risk-aware safe set $\mathcal{C}_b$ defined over beliefs, find a control input $\bm{u}$ that renders $\mathcal{C}_b$ forward invariant.
\end{problem}
By solving Problem \ref{prob:Belief}, we ensure  that the belief satisfies a $\mathrm{VaR}$ formulation which is qualitatively equivalent (see Eq.~\eqref{eq:qual}) to satisfying the chance constraint in Problem \ref{prob:State}.
\begin{example}
Consider a drone operating in one dimension with position $x \sim \mathcal{N}\left(\mu, \sigma^2\right)$, as shown in Fig. \ref{fig:Example} and the Gaussian belief state $\bm{b} = \left[\mu, \sigma^2\right]^T \in \mathbb{R}^2$. A safety specification over states is to stay within the collision-free space with 90\% probability, given by  $\delta = 0.1$ 
and $\mathcal{C}_x = \left\{ x\in \mathbb{R} \mid x \leq 2\right\}$. The corresponding safe set $\mathcal{C}_b$ in belief space is defined through
\begin{align*}
    h\left(\bm{b}\right) &= \mathrm{VaR}_{\delta}\left(2 - x\right) = 2 - \mu + \mathrm{erf}^{-1}(1-2\delta) \sqrt{2 \sigma^2} \geq 0,
\end{align*}
which is illustrated in blue in Fig. \ref{fig:Example}. In the depicted point in time, the belief state, shown as red circle, is right at the boundary of the safe set $\partial \mathcal{C}_b$ as the probability of colliding with the wall is exactly 10~\%. Solving Problem \ref{prob:Belief} keeps the belief state (red circle) in $\mathcal{C}_b$ which in turn satisfies the original safety specification over states that $90\%$ of the drone's probability mass should be left of the wall.
\end{example}

\section{Risk-aware Control}
Our solution to Problem \ref{prob:Belief} is divided into two main steps. First, we derive a stochastic hybrid system that accounts for uncertainties in the robot's motion and observations. Given this hybrid system describing the evolution of the robot's belief, we propose BCBFs to ensure forward invariance of our safe set defined over beliefs. BCBFs serve as a computationally efficient risk-aware safety filter that can be applied to various robotic systems under stochastic uncertainties.

\subsection{Hybrid Belief Dynamics}
\label{sec:BeliefDynamics}
In the belief dynamics, it is important to distinguish between two  evolutions: 1) The belief advances in continuous time \high{as the state dynamics~\eqref{eq:motion} also evolve in continuous time}. 2) At discrete timesteps $t_k$ when a new sensor observation is available, the belief changes instantaneously since the state distribution is conditioned on the measurement. This naturally leads to a hybrid dynamical system which is derived from a continuous-discrete EKF~\cite{lewisbook, nishimura2021sacbp}. 
\subsubsection{Continuous-time Belief Dynamics}
The continuous-time evolution of the belief directly follows from the derivation of the continuous EKF and is given by a set of ordinary differential equations (ODEs)
\begin{align}
\dot{\bm{\mu}} &= \bm{f}\left(\bm{\mu}\right) + \bm{g}\left(\bm{\mu}\right) \bm{u}, \hspace{0.3cm} \label{eq:mean}\\
\dot{\bm{\Sigma}} &= \bm{A} \bm{\Sigma} + \bm{\Sigma} \bm{A}^T + \bm{Q},\label{eq:cov}
\end{align}
where $\bm{A}$
is the Jacobian of the noise-free motion model in Eq.~\eqref{eq:motion} evaluated at the current mean $\bm{\mu}$. For the design of BCBFs, the dynamics are required to be in control-affine form which has been proven to be true in \cite{nishimura2021sacbp}.

Since the belief vector $\bm{b}$ is comprised of the mean and the covariance matrix, we can express the continuous-time belief dynamics as a vector-valued ODE
\begin{align}
    \dot{\bm{b}} = \begin{bmatrix}
    \dot{\bm{\mu}}\\
    \mathrm{vec} \left(\dot{\bm{\Sigma}}\right)
    \end{bmatrix} = \underbrace{\begin{bmatrix}
    \bm{f}\left(\bm{\mu}\right)\\
    \bm{f}_{\Sigma}\left(\bm{b}\right)
    \end{bmatrix}}_{\bm{f}_b\left(\bm{b}\right)} + \underbrace{\begin{bmatrix}
    \bm{g}\left(\bm{\mu}\right)\\
    \bm{g}_{\Sigma}\left(\bm{b}\right)
    \end{bmatrix}}_{\bm{g}_b\left(\bm{b}\right)} \bm{u}, \label{eq:conttime}
\end{align}
where $\bm{f}_{\Sigma} = \mathrm{vec}\left(\bm{F}_{\Sigma}\right)$ and $\bm{g}_{\Sigma} = \mathrm{vec}\left(\bm{G}_{\Sigma}\right)$ are the vector fields of Eq.~\eqref{eq:cov} in control-affine form. Note that $\bm{G}_{\Sigma} \in \mathbb{R}^{n \times n \times m}$ and, thus, the $\mathrm{vec}(\cdot)$ operator maps it to a matrix $\bm{g}_{\Sigma} \in \mathbb{R}^{\nicefrac{n(n+1)}{2} \times m}$. 

Consequently, if the control signal $\bm{u}$ is known, the belief at any point in time is obtained by forward integrating the belief dynamics in Eq. \eqref{eq:conttime}. However, when a new sensor reading is available, the belief changes instantaneously which is covered in the discrete-time Kalman update.

\subsubsection{Discrete-time Kalman Update}
Every time a new observation $\bm{z}_k$ is available, we obtain the posterior distribution $p\left(\bm{x}\left(t_k^+\right) \mid \bm{z}_{1:k-1}, \bm{z}_k\right) = \mathcal{N}\left(\bm{\mu}^+, \bm{\Sigma}^+ \right)$ by conditioning the prior distribution parameterized by $\bm{b}^- = \bm{b}\left(t_k^-\right)$ on $\bm{z}_k$. 
Note that $t_k^-$ is infinitesimal smaller than $t_k = t_k^+$. 
The conditioning leads to a discrete belief transition which is governed by the discrete-time Kalman update \cite{thrun2005probabilistic}
\begin{align}
    \bm{b}^+ = \bm{\Delta}\left(\bm{b}^-\right) = \begin{bmatrix}
    \bm{\mu}^- + \bm{K} \left(\bm{z}_k - \bm{\ell}\left(\bm{\mu}^-\right)\right)\\
    \mathrm{vec} \left(\left(\bm{I} - \bm{K} \bm{H}\right) \bm{\Sigma}^- \right)
    \end{bmatrix}, \text{ where}\label{eq:DiscreteTransition}
\end{align}
\begin{align*}
    \bm{K} = \bm{\Sigma}^- \bm{H}^T \left(\bm{H} \bm{\Sigma}^- \bm{H}^T + \bm{R}\right)^{-1} \text{ and }
    \bm{H} = \frac{\partial \bm{\ell}}{\partial \bm{x}} \Big|_{\bm{\mu}^-}
\end{align*}
denote the Kalman gain and the Jacobian of the observation model, respectively. The measurement obtained at a discrete timestep is not known in advance which makes it generally difficult to do planning or control in belief space. However, the unknown measurement can be treated as a random variable making the measurement update \eqref{eq:DiscreteTransition} stochastic \cite{van2012motion}.
\begin{proposition}
\label{prop:Innovation}
Under Assumption \ref{assumption:linearizable}, the innovation term $\bm{\theta} = \bm{K} \left(\bm{z}_k - \bm{\ell}\left(\bm{\mu}^-\right)\right)$ in Eq. \eqref{eq:DiscreteTransition} is a random variable with distribution $\bm{\theta} \sim \mathcal{N} \left(\bm{0}, \bm{\Lambda}\right)$ where
\begin{align}
    \bm{\Lambda} = \bm{K} \left(\bm{H} \bm{\Sigma} \bm{H}^T + \bm{R}\right)\bm{K}^T.
\end{align}
\end{proposition}%
The proof is straight-forward by applying Assumption \ref{assumption:linearizable} to the innovation term and is omitted for brevity. 

Finally, combining the ODE in Eq. \eqref{eq:conttime} and the discrete-time Kalman update in Eq. \eqref{eq:DiscreteTransition} leads to the hybrid system
\begin{align}
    \mathcal{S} = 
    \begin{cases}
        \dot{\bm{b}} = \bm{f}_b\left(\bm{b}\right) + \bm{g}_b\left(\bm{b}\right) \bm{u}, & \forall t \in [t_{k-1}, t_k)\\
        \bm{b}^+ = \bm{\Delta}\left(\bm{b}^-\right) & t = t_k,
    \end{cases}\label{eq:hybridsys}
\end{align}
describing the evolution of the robot's belief over time. Fig. \ref{fig:safeset} illustrates an example trajectory of the belief. 
Our formulation has the advantage that we can allow arbitrary sampling times $\delta t = t_k - t_{k-1}$. This is particularly useful for the analysis of scenarios where e.g., a sensor does not provide any information for a certain duration as in the case of GPS data inside a building.

\subsection{Belief Control Barrier Functions}
The dynamical system $\mathcal{S}$ of the belief allows us to introduce BCBFs for safe control under stochastic uncertainties. \high{BCBFs are defined similarly as CBFs (\ref{def:CBF}),  but they are defined over beliefs instead of states.} 
\begin{definition}
(BCBF) Given a safe set $\mathcal{C}_b$ defined by Eq.~\eqref{eq:safeset}, $h_b\left(\bm{b}\right)$ serves as a Belief Control Barrier Function (BCBF) for the stochastic dynamical system \eqref{eq:motion}-\eqref{eq:observation} if  $\forall \bm{b}$ satisfying $h_b\left(\bm{b}\right) \geq 0, \exists \bm{u} \in \mathcal{U}$ such that
\begin{align}
    \frac{\partial h_b}{\partial \bm{b}} \left(\bm{f}_b\left(\bm{b}\right) + \bm{g}_b\left(\bm{b}\right) \bm{u}\right) \geq - h_b\left(\bm{b}\right).\label{def:BCBF}
\end{align}
\end{definition}
Using this definition of a BCBF, the following theorem provides a condition for the forward invariance of the safe set $\mathcal{C}_b$ under the continuous-time evolution of the belief.
\begin{theorem}
\label{thm:ContInvariance}
If a locally Lipschitz control input $\bm{u}(t)$ satisfies Eq. \eqref{def:BCBF} $\forall t \in [t_k, t_{k+1})$ for a given safe set \high{with valid BCBF $h_b(\bm{b})$}, then $\mathrm{Pr}\left[h_b\left(\bm{b}(t)\right) \geq 0 \right] = 1$, provided that $\bm{b}(t_k) \in \mathcal{C}_b$.
\end{theorem}
The proof is straight-forward since we only reason about the continuous-time evolution in the interval $[t_k, t_{k+1})$ in which the belief dynamics in Eq. \eqref{eq:conttime} are deterministic. Thus, interested readers are referred to the proof for traditional state-based CBFs \cite{ames2016control}. 

\begin{figure}[t]
    \centering
    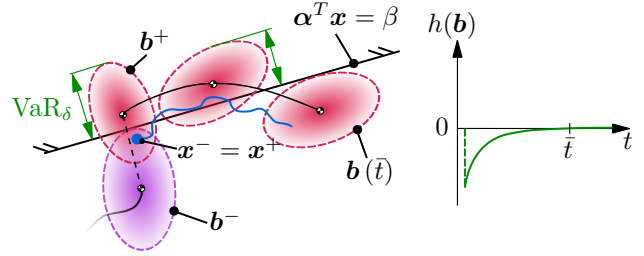
    \caption{An example belief trajectory subject to an outlier measurement and the corresponding safe set shown as half-space. The initial belief $\bm{b}^-$ is within the safe set $\mathcal{C}_b$ whereas the posterior belief $\bm{b}^+$ leaves the safe set under the discrete transition. An asymptotically stable belief trajectory is shown in red while the true evolution of the state is shown in blue. The plot on the right shows value of $h_b\left(\bm{b}\right)$, depicted in green.}
    \vspace{-0.5cm}
    \label{fig:safeset}
\end{figure}

Next, we analyze the effects of measurements on the forward invariance of $\mathcal{C}_b$. Since measurements are treated as random variables, we do not have control of the discrete belief transition.
As a result, outlier measurements may cause the belief to leave the safe set $\mathcal{C}_b$ even though the considered state is safe, see Fig.~\ref{fig:safeset}.
We show how to ensure forward invariance even in the presence of outlier measurements. 
To that end, we exploit the asymptotic stability of zeroing CBFs.
\begin{remark}
\label{remark:stable}
Our BCBFs are zeroing CBFs that not only render $\mathcal{C}_b$ forward invariant but also asymptotically stable \high{as the cont.-time belief dynamics are deterministic} \cite{ames2019control}. Thus, if an initial belief is outside the safe set, it will be driven back to $\mathcal{C}_b$ over time.
\end{remark}

In the following theorem, we derive an upper bound on the probability of leaving $\mathcal{C}_b$ under a discrete transition.
\begin{theorem}
\label{thm:NaturalBound}
    If the control input $\bm{u}(t)$ satisfies Eq. \eqref{def:BCBF}, the probability of leaving the safe set under a discrete transition, i.e. $\mathrm{Pr}\left[h_b\left(\bm{b}^+\right) < 0\right]$, is bounded by
    \begin{align}
        \mathrm{Pr}\left[h_b\left(\bm{b}^+\right) < 0\right] \leq
     \frac{1}{2}\left(1 - \mathrm{erf}\left(\frac{\xi\left(\bm{b}^-\right)}{\sqrt{2 \bm{\alpha}^T \bm{\Lambda} \bm{\alpha}}}\right)\right),
    \end{align}
    with $\bm{\Lambda}$ defined as in Proposition \ref{prop:Innovation} and
    \begin{align}
        \xi\left(\bm{b}^-\right) &= \mathrm{erf}^{-1} \left(1 - 2 \delta \right)\bigg(\sqrt{2 \bm{\alpha}^T \bm{\Sigma^-}\bm{\alpha}}\\&
        - \sqrt{2 \bm{\alpha}^T \bm{\left(\bm{I} - \bm{KH}\right)\Sigma^-}\bm{\alpha}}\bigg).
    \end{align}
\end{theorem}
\emph{Proof:}
Since $\bm{u}$ satisfies Eq. \eqref{def:BCBF} we know that $h_b\left(\bm{b}^-\right)=h_b\left(\bm{b}\left(t_k^-\right)\right) \geq 0$. Consequently, the value of the BCBF is
 \begin{align*}
    h_b\left(\bm{b}^+\right) &= \bm{\alpha}^T \bm{\mu^+} - \beta- \mathrm{erf}^{-1}(1-2\delta) \sqrt{2 \bm{\alpha}^T \bm{\Sigma^+}\bm{\alpha}}
\end{align*}
in which we replace the posterior mean $\bm{\mu}^+$ by its stochastic update equation from Proposition \ref{prop:Innovation}. Further, we add a covariance term that adds up to zero,
\begin{align*}
     h_b\left(\bm{b}^+\right) &=\bm{\alpha}^T \left(\bm{\mu^-} + \bm{\theta}\right) - \beta -\mathrm{erf}^{-1}(1-2\delta) \sqrt{2 \bm{\alpha}^T \bm{\Sigma^+}\bm{\alpha}}\\
    &+ \mathrm{erf}^{-1}(1-2\delta) \left(\sqrt{2 \bm{\alpha}^T \bm{\Sigma^-}\bm{\alpha}} - \sqrt{2 \bm{\alpha}^T \bm{\Sigma^-}\bm{\alpha}}\right).
\end{align*}
which we rearrange to obtain $h_b\left(\bm{b}^-\right)$. This yields a Gaussian distribution over possible BCBF values
\begin{align}
    h_b\left(\bm{b}^+\right) &= h_b\left(\bm{b}^-\right) + \bm{\alpha}^T \bm{\theta} + \xi\left(\bm{b}^-\right). \label{eq:hjump}
\end{align}
Finally, the probability that an outlier measurement causes the belief to leave $\mathcal{C}_b$ is calculated using \eqref{eq:ProbHalfspace} as
\begin{equation*}
\begin{aligned}
\mathrm{Pr}\left[h_b\left(\bm{b}^+\right) < 0\right] &= \mathrm{Pr}\left[- \bm{\alpha}^T \bm{\theta} \geq h_b\left(\bm{b}^-\right) + \xi\left(\bm{b}^-\right)\right]\\
&= \frac{1}{2}\left(1 - \mathrm{erf}\left(\frac{\xi\left(\bm{b}^-\right) + h_b\left(\bm{b}^-\right)}{\sqrt{2 \bm{\alpha}^T \bm{\Lambda} \bm{\alpha}}}\right)\right)\\
&\leq \frac{1}{2}\left(1 - \mathrm{erf}\left(\frac{\xi\left(\bm{b}^-\right) }{\sqrt{2 \bm{\alpha}^T \bm{\Lambda} \bm{\alpha}}}\right)\right)\hspace{0.7cm}\blacksquare
\end{aligned}
\end{equation*}
Theorem \ref{thm:NaturalBound} provides us with a natural bound on the probability of leaving the safe set $\mathcal{C}_b$ under discrete sensor observations when the belief state $\bm{b}^-$ is on the  boundary $\partial \mathcal{C}_b$. 
 The derived natural bound only depends on our observation model $\bm{\ell}\left(\bm{x}\right)$, the noise covariance $\bm{R}$ and the prior covariance $\bm{\Sigma^-}$. Thus, the specific bound varies for different robotic systems and sensor types.
\begin{example}
(Cont.) Consider the one-dimensional drone example introduced before. The drone is equipped with a sensor that provides noisy state observations $z_k = x(t_k) + v$ where $v \sim \mathcal{N}\left(0, r=0.1\right)$. The prior variance is $\sigma^{-} = 0.3$ and a confidence level is set to $1 - \delta = 0.99$, then
\begin{align*}
     &\mathrm{Pr}\left[h_b\left(\bm{b}^+\right) < 0\right] \leq\\ &\frac{1}{2}\left(1 - \mathrm{erf}\left(\mathrm{erf}^{-1}\left(1 - 2\delta\right) \left(\frac{\sqrt{\sigma^- + r} - \sqrt{r}}{\sqrt{\sigma^-}} \right) \right)\right) \approx 0.09
\end{align*}
meaning that the probability of leaving $\mathcal{C}_b$ is at most 9 \%. Note that this is a tight bound if the prior belief state is right at the boundary of the safe set and strictly smaller otherwise.
\end{example}
Figure \ref{fig:safeset} shows a scenario in which the posterior belief $\bm{b}^+$ leaves the safe set due to an outlier measurement. 
Although the state $\bm{x}$ still satisfies the original half-space constraint at time $t_k$, the belief switches instantaneously from $h_b\left(\bm{b}^-\right) \geq 0$ to $h_b\left(\bm{b}^+\right) < 0$. This highlights the importance of keeping the belief state inside $\mathcal{C}_b$: Initially, the state $\bm{x}$ is considered to be safe, i.e. $\bm{x} \in \mathcal{C}_x$, but as time evolves, the only statement we can make is that the belief trajectory $\bm{b}(t)$ is asymptotically stable, see Remark~\ref{remark:stable}. \high{However, during this period $\bm{x}$ could leave $\mathcal{C}_x$ while the belief is driven back to $\mathcal{C}_b$ as shown in Fig.~\ref{fig:safeset}. We thus need to ensure that the belief state does not leave $\mathcal{C}_b$ in the first place, in order to guarantee that $\bm{x}$ remains in $\mathcal{C}_x$ with desired probability.} 



To ensure that the belief state does not leave $\mathcal{C}_b$, we modify $\mathcal{C}_b$ such that the probability of leaving $\mathcal{C}_b$ is bounded with a desired confidence. We define an augmented safe set $\tilde{\mathcal{C}}_b \subseteq \mathcal{C}_b$
\begin{align}
    \tilde{\mathcal{C}}_b = \left\{\bm{b} \in \mathbb{R}^{n_b} \mid 
   \tilde{h}_b(\bm{b}) \geq 0 \right\}, \label{eq:safeset_tilde}
\end{align}
with $\tilde{h}_b\left(\bm{b}\right) = h_b\left(\bm{b}\right) - \gamma$ for some $\gamma \geq 0$. This safe set is essentially a shrunk version of the original safe set $\mathcal{C}_b$. In the following theorem, we choose $\gamma$ such that the belief stays within $\mathcal{C}_b$ with desired probability.
\begin{theorem}
    \label{thm:AugmentedSafeset} 
    The belief state $\bm{b}^+$ remains in the safe set $\mathcal{C}_b$ with probability $\mathrm{Pr}\left[\bm{b}^+ \in \mathcal{C}_b\right] \geq 1 - \varepsilon$ under the discrete reset map $\bm{b}^+ = \bm{\Delta}\left(\bm{b}^-\right)$  if the control input $\bm{u}(t)$ satisfies Eq.~\eqref{def:BCBF} for the \high{function $\tilde{h}_b\left(\bm{b}\right)$ (thus $\tilde{h}_b$ is a valid BCBF)} and $\tilde{h}_b\left(\bm{b}^-\right) \geq 0$ for %
    \begin{align*}
        \gamma &\geq  \sqrt{2 \bm{\alpha}^T \bm{\Lambda}\bm{\alpha}}\left(\mathrm{erf}^{-1}(1-2\varepsilon)\right) - \xi\left(\bm{b}^-\right).
    \end{align*}
\end{theorem}
\emph{Proof:} Similarly as for Th.~\ref{thm:NaturalBound}, we calculate the probability of staying in the safe set $\mathcal{C}_b$ using Eq. \eqref{eq:hjump},
\begin{align*}
    \mathrm{Pr}\left[h\left(\bm{b}^+\right) \geq 0\right] &= \frac{1}{2}\left(1 + \mathrm{erf}\left(\frac{\xi\left(\bm{b}^-\right) + h\left(\bm{b}^-\right)}{\sqrt{2 \bm{a}^T \bm{\Lambda} \bm{a}}}\right)\right)\\
    &\geq \frac{1}{2}\left(1 + \mathrm{erf}\left(\frac{\xi\left(\bm{b}^-\right) + \gamma}{\sqrt{2 \bm{a}^T \bm{\Lambda} \bm{a}}}\right)\right)= 1 - \varepsilon. \hspace{0.1cm}\blacksquare
\end{align*}
Note that Theorem~\ref{thm:AugmentedSafeset} provides probabilistic safety guarantees for the case that $\bm{b}^- \in \tilde{\mathcal{C}}_b$.
If the posterior belief state ends up in the set $\mathcal{C}_b \setminus \tilde{\mathcal{C}}_b$, it will be driven back to the augmented safe set \high{if $\tilde{h}_b$ is a valid BCBF}.
If an outlier measurement occurs during this asymptotically stable convergence to $\tilde{\mathcal{C}}_b$, the natural bound in Theorem~\ref{thm:NaturalBound} holds.
Once $\bm{b}\in\tilde{\mathcal{C}}_b$, the desired confidence in Theorem~\ref{thm:AugmentedSafeset} holds again.

\subsection{Risk-aware control synthesis}
To synthesize risk-aware control inputs under stochastic uncertainties arising from state estimation, we formulate the quadratic program (QP)
\begin{equation}
\begin{aligned}
    \bm{u}^* = \argmin_{\bm{u} \in \mathcal{U}} \quad & \left(\bm{u} - \bm{u}_{\text{ref}}\right)^T \left(\bm{u} - \bm{u}_{\text{ref}}\right)\\
    \textrm{s.t.~~} \quad & \frac{\partial \tilde{h}}{\partial \bm{b}} \left(\bm{f}_b\left(\bm{b}\right) + \bm{g}_b\left(\bm{b}\right) \bm{u}\right) \geq - \tilde{h}\left(\bm{b}\right),
\end{aligned}
\label{eq:QP}
\end{equation}
where $\bm{u}_{\text{ref}}$ is a reference control input. This synthesis problem can be solved efficiently using off-the-shelf QP solvers. 

Consider for example a quadrotor operating in 3D: It has 6 degrees of freedom (3D pose and orientation) and 4 control inputs (thrusters) which leads to a 12 dimensional state. Consequently, the belief state is of dimension $\bm{b} \in \mathbb{R}^{90}$ which \high{is known as the curse of dimenionality} and complicates the use of common optimization-based controllers such as MPC. \high{We overcome the dimenionality problem of belief spaces by only optimizing over control inputs $\bm{u}$ which enables real-time applicability. The resulting control inputs satisfy the entire belief dynamics.}

\section{Experiments}
We evaluate our BCBFs in challenging scenarios in which uncertainties cannot be neglected\footnote{Videos of our hardware experiments can be found in the supplementary material}.
Specifically, we show
\begin{enumerate}
    \item improved adherence to safety specifications through risk-awareness and computational efficiency over traditional CBFs and CCNMPC in simulations in Sec.~\ref{sec:sim},
    \item safety under changing sensing conditions, such as sensor rates and measurement variances, in Sec.~\ref{exp:corridor}, 
    \item robustness to external disturbances as well as the generality to different sensor systems in Sec.~\ref{exp:disturbances}.
\end{enumerate}

\subsection{Safety analysis of BCBFs}
\label{sec:sim}

\paragraph{Setup}
We compare our BCBFs to two baselines in an obstacle avoidance scenario. We consider a unicycle robot \cite{cortez2021robust} with state $\bm{x} = [p_x, p_y, v, \varphi]^T$ and dynamics and observations
\begin{equation}
\begin{aligned}
    \dot{p}_x &= v ~ \mathrm{cos} \left(\varphi\right) + w_x, \hspace{0.5cm} \dot{v} = a +w_v\\
    \dot{p}_y &= v ~ \mathrm{sin} \left(\varphi\right) + w_y, \hspace{0.5cm} \dot{\varphi} = \omega + w_{\varphi}\\
    \bm{z}_k &= \bm{x}_k + \bm{v}_k, \bm{v}_k \sim \mathcal{N}\left(\bm{0}, \bm{R})\right).
\end{aligned}
\label{eq:Unicycle}
\end{equation}
\begin{figure}[t]
    \centering
    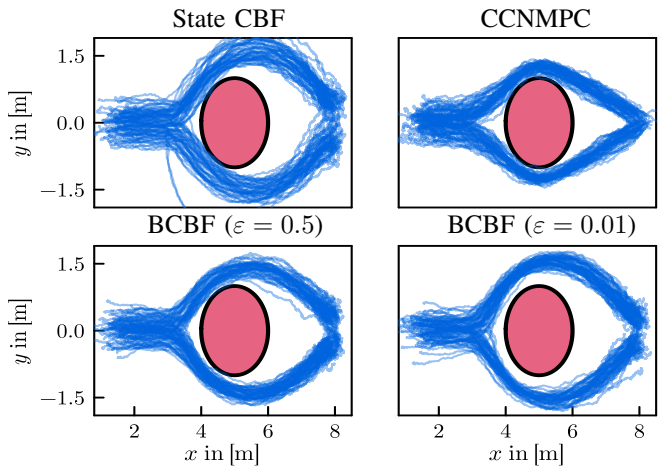
    \caption{Comparison of different control strategies for a two-dimensional avoidance scenario. The initial state follows a Gaussian distribution $\bm{x}_0 \sim \mathcal{N}\left([0,1.5, 0, 0], \mathrm{diag}\left([0.3^2, 0.2^2, 0.1^2, 0.1^2]\right)\right)$ and the goal is at [8, 0]. Simulated trajectories are shown in blue and the obstacle is depicted in red.}
    \label{fig:SimUnicycle}
    \vspace{-0.6cm}
\end{figure}
The nominal dynamics are corrupted by Gaussian noise $[w_x, w_y, w_v, w_{\varphi}]^T \sim \mathcal{N}\left(\bm{0}, \bm{Q}\right)$ with motion noise $\bm{Q} = \mathrm{diag}\left(\left[0.1^2, 0.1^2, 0.005^2, 0.005^2\right]\right)$ and observation noise $\bm{R} = \mathrm{diag}([0.2^2, 0.2^2, 0.1^2, 0.1^2])$ with a sensor update rate of 10~Hz. The control input is given as $\bm{u}=[a, \omega]^T$. 
We obtain the belief dynamics in Eq. \eqref{eq:hybridsys} by applying the method described in Sec.~\ref{sec:BeliefDynamics} where $\bm{b} \in \mathbb{R}^{14}$. The objective is to steer the robot towards a goal at $[8,0]$ while avoiding collisions with a circular obstacle $\mathcal{O}=\left\{\bm{p} \in \mathbb{R}^2\mid \lVert \bm{p} - \bm{c}\rVert_2 < r\right\}$ where $\bm{c} = [5,0]^T$ and $r = 1$ as shown in Fig. \ref{fig:SimUnicycle}. 
The safety specification for collision avoidance is given as
\begin{align}
    \mathrm{Pr}\left[\bm{x} \notin \mathcal{O} \right] = \mathrm{Pr}\left[\lVert \bm{x} - \bm{c}\rVert_2 -r \geq 0\right] \geq 1 - \delta \label{eq:CollCond}
\end{align} 
with $\delta = 0.01$. 
Due to the non-linearity of the obstacle description, we use the  common approach to linearize Eq. \eqref{eq:CollCond} around the mean state which is a strict overapproximation of the true collision condition as shown in \cite{8613928}. The resulting risk-aware safe set is given by
\begin{align}
    \mathcal{C}_b = \left\{\bm{b} \in \mathbb{R}^{14} \mid \mathrm{VaR}_{\delta}\left(\bm{\alpha}^T \left(\bm{x} - \bm{c}\right) - r \right) \geq 0\right\}\label{eq:simsafeset}
\end{align}
where \high{$\mathrm{VaR}_{\delta}(\cdot)$ with $\bm{\alpha} = \nicefrac{(\bm{\mu} - \bm{c})}{\lVert \bm{\mu} - \bm{c} \rVert_2}$ serves as a BCBF candidate}. Due to a relative degree $r_b=2$ in our \high{BCBF candidate}, we use second order formulations as in Eq.\eqref{eq:SecondOrderCBF}.

We solve the resulting SDE \eqref{eq:Unicycle} using the DifferentialEquations.jl package \cite{rackauckas2017differentialequations} in the Julia programming language. 
The reference control $\bm{u}_\textrm{ref}$ is given by a Linear-quadratic Regulator (LQR) that uses the mean estimate $\bm{\mu}$ to steer the system towards the goal without any knowledge of the obstacle. \high{The parameters chosen for the LQR are $\bm{Q}_L = \textrm{diag}[10,5,5,5]$ and $\bm{R}_L=\textrm{diag}[5,10]$.}
We compare two versions of BCBFs, namely one with only the natural bound ($\varepsilon=0.5$) and one bounding the probability of leaving $\mathcal{C}_b$ by $\varepsilon=0.01$.

\paragraph{Baselines}
We compare our BCBFs to two different baselines: \high{1.) Stochastic CBFs with incomplete state information which require cont.-time observations as well as a bounded estimation error. To enable continuous estimation, we use maximum likelihood observations between actual measurements which is not true in practice. The bounded estimation error was obtained by running a monte carlo study and take the maximum error.} 
2.) chance-constrained NMPC (CCNMPC) \cite{8613928} that formulates an MPC problem which includes Eq.~\eqref{eq:simsafeset} as a constraint. Since MPC operates in discrete-time while we simulate a continuous system, we use a zero order hold to apply the control input. We run the MPC at 30~Hz and use a planning horizon of $N=40$. The cost function is the same quadratic cost as in the reference LQR controller. \high{3.) In hardware experiments, we compare to a traditional state CBF that only considers the mean dynamics.}

\paragraph{Safety and computational efficiency}
Figure \ref{fig:SimUnicycle} shows 100 simulated trajectories for different initial conditions sampled from a Gaussian while the quantitative results are summarized in Table \ref{tab:simresults}. 
BCBFs have the lowest number of collisions (0 for $\varepsilon=0.01)$ and outperform both baselines.
\high{Interestingly, SCBFs cannot ensure safety almost surely as the assumption of cont.-time observations does not apply.} Similarly, CCNMPC results in multiple collisions although it tries to satisfy the same safety specification. 
This is due to the fact that the optimization problem sometimes becomes infeasible in which case we need to relax the safety constraint.
In \cite{8613928}, the authors reported that the percentage of infeasible solutions in their experiments was $2.8 \%$. 
Infeasibility is not an issue for BCBFs, since they allow to formulate the control synthesis as QP over controls instead of a nonlinear program over states and controls, overcoming the curse of dimensionality. 
QP formulations are computationally efficient and can be run with up to 1kHz, see Table \ref{tab:simresults}. 

\paragraph{Effect of augmented safe set}
We analyze the effect of the augmented safe set defined in Theorem \ref{thm:AugmentedSafeset}. If no additional safety margin is used ($\varepsilon=0.5$), the belief leaves $\mathcal{C}_b$ in $0.42 \%$ of all simulated trajectories. 
When bounding the probability of leaving the safe set by $\mathrm{Pr}[h\left(\bm{b}^+\right) < 0] \leq 0.01$, the belief remains in $\mathcal{C}_b$ for all trajectories while keeping larger distances to the obstacle. 

{\renewcommand{\arraystretch}{1.15}
\begin{table}[t]
    \caption{Comparison of BCBFs to baselines.}
    \resizebox{0.47\textwidth}{!}{\begin{tabular}{l | cccc}
                   & \high{\begin{tabular}[c]{@{}c@{}}\bf{Stochastic} \\ \bf{CBF}\end{tabular}} & \bf{CCNMPC} & \begin{tabular}[c]{@{}c@{}}\bf{BCBF} \\ ($\varepsilon = 0.5$)\end{tabular} & \begin{tabular}[c]{@{}c@{}}\bf{BCBF} \\ ($\varepsilon = 0.01$)\end{tabular} \\
                   \hline\hline
    \# collisions  &     \high{3}      &   3    &      1    &      \bf{0}   \\
    \hline
    \% $h(\bm{b}) < 0$ &     \high{6.52}      &    0.78   &      0.42    & \bf{0}\\
    \hline
    \begin{tabular}[c]{@{}c@{}}Avg. $t_{\text{comp}}$  in [s] \\ $(\mu \pm \sigma)$\end{tabular}   &      \high{\begin{tabular}[c]{@{}c@{}}$0.0008 \pm$ \\ $0.004$\end{tabular}}    &   \begin{tabular}[c]{@{}c@{}}$0.35 \pm$ \\ 0.052\end{tabular}  &    \begin{tabular}[c]{@{}c@{}}$\bm{0.0008} \pm$ \\ $\bm{0.003}$\end{tabular}     &    \begin{tabular}[c]{@{}c@{}}$0.0017 \pm$ \\ 0.0027\end{tabular}     \\
    \hline
    \begin{tabular}[c]{@{}c@{}}Avg. $\lVert\bm{u}\rVert_2$ \\ $(\mu \pm \sigma)$\end{tabular}&      \high{\begin{tabular}[c]{@{}c@{}}$1.63 \pm$ \\ 2.54\end{tabular}}      &    \begin{tabular}[c]{@{}c@{}}$\bm{0.67 \pm}$ \\ $\bm{0.4}$\end{tabular}   &    \begin{tabular}[c]{@{}c@{}}$1.62 \pm$ \\ 1.65\end{tabular}      &   \begin{tabular}[c]{@{}c@{}}$1.52 \pm$ \\ 1.6\end{tabular}\\
    \hline
    \begin{tabular}[c]{@{}c@{}}\high{$t_{g}$ in [s]} \\ \high{$(\mu \pm \sigma)$}\end{tabular} & \high{\begin{tabular}[c]{@{}c@{}}$4.01 \pm$ \\ 0.5\end{tabular}} & \high{\begin{tabular}[c]{@{}c@{}}$3.35 \pm$ \\ 0.19\end{tabular}} & \high{\begin{tabular}[c]{@{}c@{}}$4.1 \pm$ \\ 0.41\end{tabular}} & \high{\begin{tabular}[c]{@{}c@{}}$4.16 \pm$ \\ 0.44\end{tabular}}
    \end{tabular}}
    \vspace{-0.6cm}
    \label{tab:simresults}
\end{table}
}

\paragraph{Control efficiency}
Lastly, we evaluate the efficiency in terms of the average norm of the control input \high{and the time $t_g$ to reach the goal}. In this case CCNMPC is superior to all other approaches since it optimizes controls over a planning horizon whereas CBF approaches are purely reactive. This motivates to combine state-based MPC with BCBFs to achieve both, control efficiency and rigorous safety properties. In future work, we aim to explore this.

\subsection{Setup of hardware experiment}

We use the Bitcraze Crazyflie 2.1 quadrotor inside an OptiTrack motion capture (mocap) system, as shown in Fig. \ref{fig:RealworldExp}. 
The state of the drone is given as its 3D position $\bm{p}$ and velocity $\dot{\bm{p}}$. Its motion and observations are modeled as
\begin{align}
    \dot{\bm{x}} &= \begin{bmatrix}
    \dot{\bm{p}}\\
    \ddot{\bm{p}}
    \end{bmatrix} = \begin{bmatrix}
    \bm{0} & \bm{I}\\
    \bm{0} & \bm{0}
    \end{bmatrix}\begin{bmatrix}
    \bm{p}\\
    \dot{\bm{p}}
    \end{bmatrix}  + \begin{bmatrix}
    \bm{0}\\
    \bm{I}
    \end{bmatrix} \bm{u} + \bm{w},\label{eq:motionexp}\\
    \bm{z}_k &= \bm{p}_k + \bm{v}_k\label{eq:obsexp}
\end{align}
where $\bm{u} \in\mathbb{R}^3$ are the drone's desired accelerations and $\bm{w} \sim \mathcal{N}\left(\bm{0}, \bm{Q}\right)$ with a variance of $0.05^2$ along the diagonal of $\bm{Q}$. 
In our setting, we can only observe the drone's position $\bm{p}$.
We add Gaussian noise $\bm{v}\sim\mathcal{N}\left(\bm{0}, \bm{R}\right)$ to the mocap data and reduce the sampling rate of the measurements. 
These parameters $\bm{R}$ and $f_s$ are varied throughout the experiments. 
We use the uncorrupted mocap data as ground truth to evaluate the performance of BCBFs.

We use the EKF in Sec.~\ref{sec:BeliefDynamics} to obtain the belief dynamics with $\bm{b} \in \mathbb{R}^{27}$, where the ODE is discretized using Euler's scheme.
We convert the desired acceleration $\bm{u}$ synthesized by our BCBFs into a setpoint consisting of desired roll and pitch angles as well as thrust for the real quadrotor using \cite{8463194}.
Setpoints are sent to the Crazyflie at 100 Hz and tracked by the \high{onboard PID controller which serves as $\bm{u}_{\text{ref}}$}.
We compare BCBFs only to traditional CBFs as we were not able to run CCNMPCs in real-time with our setup. 
\begin{figure}[t]
\centering
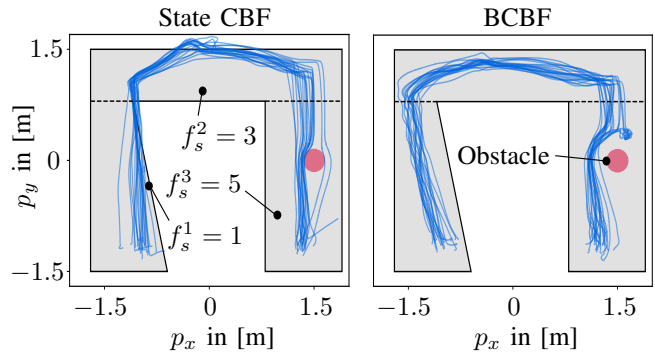
\caption{Ground truth trajectories of 20 different runs of Exp.~\RNum{1} for CBFs and BCBFs. The sampling rate $f_s$ changes throughout the corridor segments. The measurement variances are given as $\bm{R}_1 = \mathrm{diag}[0.08^2, 0.08^2]$, $\bm{R}_2 = \mathrm{diag}[0.2^2, 0.2^2]$ and $\bm{R}_3 = \mathrm{diag}[0.15^2, 0.15^2]$, respectively.}
\vspace{-0.6cm}
\label{fig:Corridor}
\end{figure}

\subsection{Experiment \RNum{1} - Changing sensing conditions}
\label{exp:corridor}
In our first experiment, we showcase the robustness of BCBFs to changing sensing conditions. 
For that purpose, we navigate the drone along a U-shaped corridor as shown in Fig. \ref{fig:Corridor}. 
The corridor is the union of three polytopes. In each of the three polytopes, we change the measurement noise $\bm{R}$ and sampling rates $f_s$ according to Fig.~\ref{fig:Corridor}. In the third polytope, we additionally place a circular obstacle to increase the difficulty of the scenario. 
The safety specification is given as an intersection of risk-aware half-spaces with a collision probability of $\delta = 0.05$. The half-spaces are given by the polytope representation of the corridor. 

Figure \ref{fig:Corridor} shows the ground truth trajectories of 20 different runs for a state-based CBF and a BCBF. It can be observed that the state-based approach cannot satisfy the safety specification whereas all 20 trajectories are collision free for the BCBF. 
Interestingly, there are no trajectories for the BCBF passing the obstacle on the right side since it is too risky to navigate through the narrow passage. 
As a consequence, the drone stops in front of the obstacle and slowly moves around the left side \high{which took about 5-10 seconds}. 
Although this behavior ensures safety, it highlights that a robot can get stuck in a local minimum when using a CBF approach. This could be overcome by combining BCBFs with a motion planner such as MPC.


\subsection{Experiment \RNum{2} - External disturbances}
\label{exp:disturbances}
In this experiment illustrated in Figure \ref{fig:RealworldExp}, we additionally study the effect of disturbances on the nominal dynamics. 
To that end, we define a safe set as a cuboid that we want the drone to stay in with 95\% probability. 
The drone is exposed to both, sensing uncertainties as well as an external wind disturbance from a fan. 
Instead of position measurements, we measure the drone's velocity through an added optical flow sensor.
These velocity measurements are given by
\begin{align*}
    \bm{z}_k = \dot{\bm{p}}_k + \bm{v}_k, \hspace{0.5cm}\bm{v} \sim \mathcal{N}\left(\bm{0}, \mathrm{diag}(0.2^2)\right).
\end{align*}
Since we \high{only measure the drone's velocity and neglect the global position measurements from mocap}, there is an inevitable drift in the position \high{estimate obtained from integrating the velocity}. 
A human operator is generating \high{reference acceleration commands $\bm{u}_{\text{ref}}$} using a gamepad and actively tries to steer the drone outside the safe set.

\begin{figure}[t]
\centering
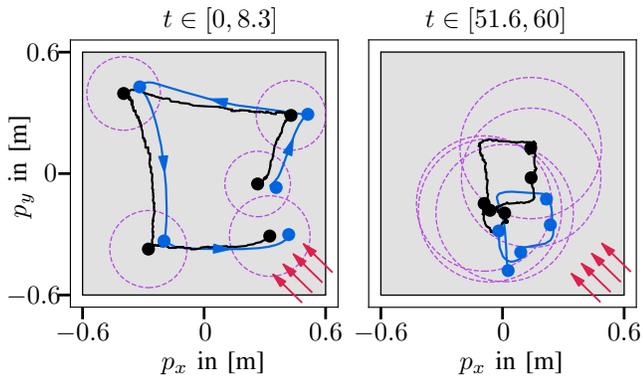
\caption{Illustration of Exp.~\RNum{2}. The drone should stay within the safety region colored in grey while a wind disturbance (shown as red arrows) is acting on the drone. The ground truth trajectory is shown in blue and the mean belief trajectory is shown in black. The 95\% confidence ellipses at selected points in time are depicted in purple.}
\vspace{-0.6cm}
\label{fig:Exp2}
\end{figure}

Figure \ref{fig:Exp2} shows the mean belief trajectory as well as ground truth for the described setting. 
The ellipses indicate the position uncertainty at selected points in time. 
The drone moves towards the corners of the safety region in which the BCBF prevents the belief from leaving the safe set. 
Over time, the uncertainty in the position estimate grows due to the lack of a global positioning system and, thus, the drone gets increasingly cautious. 
The mean belief moves further towards the center of the safety region as this increases the likelihood of satisfying the safety specification. 
At all times, the ground-truth state stays within the safety region.
\section{CONCLUSIONS AND FUTURE WORK}
Our work enables risk-aware control synthesis for stochastic dynamical systems with incomplete state information by combining continuous-discrete EKFs with CBFs defined over Gaussian belief states. Instead of defining safety specifications as hard constraints on the state, we consider a risk-aware approach in which we bound the probability of violation. 
BCBFs are applicable to any robotic system in which the state estimate is provided by an EKF. 
Our simulation and hardware experiments show that BCBFs ensure that robots adhere to safety specifications in the presence of both, real-world motion and observation uncertainties.

In future, we aim to explore the combination of motion planners such as state-based MPC that only consider the mean estimate with our proposed BCBFs. In that way, we can ensure safety while enhancing control efficiency. We are also interested in extending our results to arbitrary belief distributions that can be represented using particles filters.

\bibliographystyle{IEEEtran}
\bibliography{refs.bib}

\end{document}